\documentclass[10pt, a4paper]{article}

\usepackage[final]{lrec2026} 

\usepackage{tcolorbox}

\usepackage{listings}
\usepackage{xcolor}
\usepackage{arev}

\colorlet{punct}{red!60!black}
\definecolor{background}{HTML}{EEEEEE}
\definecolor{delim}{RGB}{20,105,176}
\colorlet{numb}{magenta!60!black}

\lstdefinelanguage{json}{
    basicstyle=\normalfont\ttfamily,
    showstringspaces=false,
    breaklines=true,
    backgroundcolor=\color{background},
    literate=
     *{0}{{{\color{numb}0}}}{1}
      {1}{{{\color{numb}1}}}{1}
      {2}{{{\color{numb}2}}}{1}
      {3}{{{\color{numb}3}}}{1}
      {4}{{{\color{numb}4}}}{1}
      {5}{{{\color{numb}5}}}{1}
      {6}{{{\color{numb}6}}}{1}
      {7}{{{\color{numb}7}}}{1}
      {8}{{{\color{numb}8}}}{1}
      {9}{{{\color{numb}9}}}{1}
      {:}{{{\color{punct}{:}}}}{1}
      {,}{{{\color{punct}{,}}}}{1}
      {\{}{{{\color{delim}{\{}}}}{1}
      {\}}{{{\color{delim}{\}}}}}{1}
      {[}{{{\color{delim}{[}}}}{1}
      {]}{{{\color{delim}{]}}}}{1},
}

\title{A Typology of Synthetic Datasets for\\ Dialogue Processing in Clinical Contexts}

\name{
Steven Bedrick$^{\clubsuit}$ \quad
A. Seza Doğruöz$^{\heartsuit}$ \quad
Sergiu Nisioi$^{\triangle}$
}

\address{
\\[0.01em]
$^{\clubsuit}$ Division of Informatics \& Clinical Epidemiology \\
\quad Oregon Health \& Science University \\[0.3em]
$^{\heartsuit}$ LT3, IDLab,  Universiteit Gent \\[0.3em]
$^{\triangle}$ HLT Research Center,  University of Bucharest \\[0.3em]
\texttt{bedricks@ohsu.edu, as.dogruoz@ugent.be, sergiu.nisioi@unibuc.ro}\thanks{All authors are corresponding authors $^{\clubsuit \heartsuit \spadesuit}$.}
\\[0.2em]
}

\abstract{
Synthetic datasets are used across linguistic domains and NLP tasks, particularly in scenarios where authentic data is limited (or even non-existent). One such domain is that of clinical (healthcare) contexts, where there exist significant and long-standing challenges (e.g., privacy, anonymization, and data governance) which have led to the development of an increasing number of synthetic datasets. 
One increasingly important category of clinical dataset is that of clinical dialogues which are especially sensitive and difficult to collect. Therefore, they are commonly synthesized.
While such synthetic datasets have been shown to be sufficient in some situations, little theory exists to inform how they may be best used and generalized to new applications. 
In this paper, we provide an overview of how synthetic datasets are created, evaluated and used for dialogue related tasks in the medical domain. Additionally, we propose a novel typology for use in classifying types and degrees of data synthesis, to facilitate comparison and evaluation.
\\ \newline }

\begin{document}

\maketitleabstract

\section{Introduction}
Synthetic datasets are increasingly used across linguistic domains and NLP tasks, particularly in scenarios where authentic data is limited (or even non-existent). One such domain is that of clinical language, where there are significant and long-standing challenges relating to privacy, anonymization, and data governance, which have led to the development of an increasing number of synthetic datasets \cite{quintana2020,long-etal-2024-llms}. 

A particularly fast-growing and relevant domain of clinical language is that of the ``clinical dialogue'' (i.e., a dialogue between a patient and a health provider, sometimes also referred to as a clinical or medical ``conversation'').
NLP tools to process such language are being rapidly adopted into practice, most notably in the form of tools for automated generation of clinical documentation~\citep{Tierney:2024,Finley:2018} as well as for more traditional dialogue applications (e.g., clinical chatbots). 
Such dialogues may occur via spoken language, or may be written (taking place via an online discussion forum, a chat-style interface, etc.).

Because of the intense commercial focus in this space, there exists a high level of need for representative dialogue datasets for system development and evaluation \citep{dogruoz-etal-2023-representativeness,adelani-etal-2025-generative}. However, due to the extremely sensitive nature of clinical conversations, genuine data is very difficult to obtain, even by the standards of clinical NLP. As such, there are a wide range of synthetic datasets purporting to represent clinical conversations of various types.

While synthetic datasets have been shown to be sufficient in some situations, little theory exists to inform scientific audiences about how these datasets may be best used and generalized to new NLP applications. Towards that end, our goals in this paper are: 1) to provide an overview of how such synthetic datasets are created, evaluated, and used for dialogue-related tasks in the medical domain via a literature review; and 2) to classify these datasets according to a novel typology.  


One of the challenges toward this goal is a lack of consensus on what is meant by the word ``synthetic'' in terms of datasets on dialogues.
The canonical definition of synthetic data is that codified by the U.S. Census Bureau in their Statistical Quality Standards (\citeyear{census:2025aa}), which may be summarized as follows: given a dataset $D$,  generate a new dataset $D'$ in a way that \textbf{a)} individual instances in $D'$ are not present in $D$, \textbf{b)} the relevant statistical properties of $D'$ match those of $D$, and \textbf{c)} original instances from $D$ may not be recovered from $D'$.\footnote{This formulation echos that proposed in the earliest literature~\citep{DonaldB.Rubin:1993} on synthetic \textit{microdata}, i.e., the individual data points contained in a dataset: sentences in a corpus, images in a collection, responses from a survey, etc.}
This definition fits well in the context of a dataset consisting of structured data. However, as noted by \citet{aggarwalGeneralSurveyPrivacyPreserving2008}, it is more challenging to apply this to textual data, for three primary reasons.

First, as compared with traditional structured data, the processes to generate novel linguistic data points are far more complex. It is relatively straightforward to generate synthetic values for scalar or categorical variables by identifying appropriate distributions and then sampling using standard numerical techniques. However, producing language that preserves the \textit{semantic} and \textit{discursive} properties of the original necessitates more sophisticated modeling and sampling techniques.
Secondly, in the context of a structured dataset, the phrase ``relevant statistical properties'' can be operationalized in terms of univariate or joint probabilities of the constituent variables. 
However, for linguistic data, the possibility space is far larger, and relying on simple statistical methods (e.g. perplexity) is unlikely to have satisfactory results. That is to say, the final dataset is unlikely to be truly ``equivalent'' to the original in linguistic terms (see \S\ref{subsec:lets_get_synthetic}).
Finally, the question of re-identifiability (constraint ``c'' from the above definition) is subtle and complex to assess in the context of linguistic data.


Another challenging aspect of synthetic data is that artificiality is not a binary property. 
All datasets, being curated and abstracted representations of some sampled reality, are inherently synthetic to some degree.
This process necessarily involves intervention around questions of inclusion/exclusion, representation, and structure. As \citet{bowkerMemoryPracticesSciences2005} put it, ``raw data is both an oxymoron and a bad idea; to the contrary, data should be cooked with care.''
However, while all datasets may be synthetic, some datasets are more synthetic than others: a dataset may be ``more'' or ``less'' synthetic, 
depending on its construction and the point of view of evaluation.
Literature on synthetic data distinguishes between datasets that are ``fully synthetic,'' being those that contain no original, genuine microdata, and ``partially synthetic,'' in which only certain variables are perturbed~\citep{reiterInferencePartiallySynthetic2003}. In terms of \textit{structured} medical data, \citet{Gonzales:2023} and \citet{Murtaza:2023} provide comprehensive discussions on this and other relevant taxonomies.

How can this framework, originally designed for numerical data, be applied to linguistic data, which focuses on meaning and grammar rather than statistics?

Exploring this question is the overall goal of our paper. In line with this goal, first, we review various ways in which synthetic language datasets are produced (\S\ref{subsec:mechanisms}). 
Next, we more formally define our terms to explore this continuum of synthesis (\S\ref{subsec:lets_get_synthetic}), and propose a typology that captures these subtleties in a practical and actionably useful way (\S\ref{subsec:typology}).
We also report the results of a literature review of datasets relating to clinical dialogues (\S\ref{subsec:synth_data_variety}) and demonstrate how our proposed typology describes existing datasets. 
Finally, we close with a discussion that extends beyond the microdata-centric view of our paper, to explore how realism and synthesis may relate to contextual aspects of a dataset (\S\ref{subsec:beyond_microdata}).

\section{Means of Production}
\label{subsec:mechanisms}

Synthetic data production methods can be categorized  into ``process-driven''\footnote{Sometimes referred to as ``theory-'' or  ``knowledge-driven.''} and ``data-driven'' methods.
Under the former, whatever generative process is carrying out the synthesis is informed by human knowledge and/or by information that is exogenous to the dataset itself. The latter describes methods that use properties of the dataset itself to derive the generative model~\citep{Goncalves:2020a,Murtaza:2023}.

This framing emphasizes a description of the mechanical process by which the synthesis is done. While important, it does not describe \emph{what} is being synthesized, nor does it address the degree to which a given dataset should be considered synthetic.
This omission is problematic when considering synthetic linguistic data, as there are numerous \textit{aspects} (e.g., topic, genre/register, grammatical or discursive structure, specific aspects of semantic content, individual lexical choices, etc.) of the data that might (or might not) be synthesized. A given dataset might involve synthesis along one of these domains, while leaving other domains un-perturbed.
To illustrate this issue, consider the question of how synthetic clinical conversational datasets are created.


From our review, we find that such datasets are created using one of several methods. First, they can be created by taking ``genuine'' data (i.e., records of authentic patient-provider conversations that actually occurred in the ``real world''), and perturbing it in some way. This perturbation may be intended to obscure identifiable information about actual patients~\citep{Negash:2023}, to produce a larger volume of comparable data than originally existed (e.g. \citet{modersohnGRASCCOFirstPublicly2022}), or as a form of data augmentation~\citep{Abdollahi:2021}.

Second, a conversational dataset can be created \textit{de novo} via a human-driven manual process. 
In medical education, it is very common for human actors to collaborate with clinical trainers or trainees to simulate various kinds of clinical interactions~\cite{Rutherford-Hemming:2024}.
Recordings of such simulated encounters, as well as other artifacts such as encounter notes, etc., can serve as rich sources of data that are free from concerns about patient privacy.
While such secondary use is rarely the \textit{goal} of simulations when done in didactic contexts, they can also be performed specifically for the purpose of generating a re-usable dataset~\citep{fareez2022dataset,yim2023aci}.

Alternatively, human annotators (trained or otherwise) may be tasked with \textit{writing} an entire end-to-end conversation given pre-existing parameters and guidelines~\citep{shi-etal-2023-midmed,ben-abacha-etal-2023-empirical}. 
In such scenarios, the resulting dialogue is not the natural result of humans interacting. It is essentially ``screenwritten,'' with the quality and utility of the resulting artifact depending heavily on the knowledge, experience, and skill of the author.

In simulated scenarios involving humans, the pragmatic and paralinguistic components of the resulting dialogues may be less representative of what would be seen in a true naturalistic exchange.
Similarly, we would expect a ``screenwritten'' scenario imagined by a single author to be less representative than a simulation involving multiple humans interacting in a more naturalistic way.
However, depending on the intended use of the resulting data, this may not be an issue. For example, if the primary content of interest is the semantic or lexical contents of the exchange, as opposed to the pragmatic or discursive contents, such a dataset may be entirely adequate. If, however, the analytical question of interest depends on discourse features, a synthetic dataset of this sort would likely be insufficiently representative.\footnote{See section~\ref{subsec:beyond_microdata} for additional discussion.} 

Finally, simulated datasets may be created \textit{de novo} via a mechanical process, e.g. using a natural language generation (NLG) model~\citep{frei2023annotated}. This may take the form of prompted generation (i.e., an LLM being prompted to write a note or simulate a dialogue given certain criteria) or in some cases by literally having two LLMs ``talk'' to one another in a simulated dialogue, in an effort to increase the variety of the resulting output~\citep{grayIncreasingRealismVariety2024}.

Human-simulated scenarios have a long history of use in medical education and evaluation~\citep{Issenberg:1999,McGaghie:2010}, resulting in a large and well-validated family of methodologies designed to ensure that the resulting interactions are sufficiently representative for their intended purpose (typically clinical training and assessment).
Despite early and longstanding interest~\citep{Starkweather:1967}, computer-driven \textit{de novo} dialogue generation has been relatively rarely used in clinical education. 
However, modern systems are beginning to be evaluated for use in medical training~\citep{Awada2024,Holderried:2024}, leading to the inevitable creation of an additional form of synthetic dataset, in which a human role-plays either the patient or the provider against the LLM playing the opposite role \cite{campillos2021lessons,abercrombie-rieser-2022-risk,zhang-etal-2023-huatuogpt} 

Given this variety of means of generation, we argue that it is inaccurate to think of a dataset as being either ``synthetic'' or ``real''. Instead, we argue that any given dataset exists as a point along a continuum between the two poles.
After all, even the most ``real-world'' datasets are inevitably the result of numerous manual interventions and decisions~\citep{gitelmanRawData2013,passiMakingDataScience2020,sambasivanEveryoneWantsModel2021}.

Similarly, we argue that an otherwise-genuine dataset that has been ``anonymized'' is, in some sense synthetic, with the degree to which this is the case depending on the method of anonymization in question. 
At one end of the spectrum, imagine a scheme in which identifiers are replaced with a class token of some kind (e.g. \texttt{PATIENT\_NAME}, \texttt{DAY\_OF\_WEEK}, etc.) in a purely deterministic and uniform manner. 
At the other end of that spectrum, consider a method in which identifiers are replaced with statistically-plausible alternative values (``pseudonymization'')\footnote{See \citet{uzunerEvaluatingStateoftheartAutomatic2007} for a detailed description of such a method, as well as an overview on deidentification in non-dialogue clinical text in general.}

In both cases, the process of applying such a method results in the creation of a new and distinct dataset.
We argue, however, that the second case is ``more synthetic'' in that there are two levels of synthesis in play: the simple alteration of the original un-modified data, in combination with some \emph{designed process} informing the pseudonymization.
This process must itself necessarily bring with it working assumptions, and may also depend on yet additional sources of data, further distancing the resulting synthetic dataset from whatever grounding in reality it may have had earlier in its development.

\begin{table*}[ht]
    \centering
    \resizebox{0.93\textwidth}{!}{
    \begin{tabular}{p{0.09\textwidth}|p{0.15\textwidth}|p{0.33\textwidth}|p{0.33\textwidth}}
        \hline
        & \textbf{Description} & \textbf{Human} & \textbf{Machine} \\
        \hline
        \textbf{Type 1}  & No intervention & - & - \\
        \hline
        \textbf{Type 2} & Altering existing microdata & Altering an existing ``real'' dialogue, according to a specification & Masking names, translating into Japanese, etc. \\
        \hline
        \textbf{Type 3} & Generating entirely new microdata & Writing a new dialogue de novo, according to a narrative prompt & LLM-generated dialogue turns \\
        \hline
    \end{tabular}%
    }
    \caption{Comparison of human and machine microdata interventions}
    \label{tab:typology}
\end{table*}

\section{What Do We Mean By ``Synthetic''?}
\label{subsec:lets_get_synthetic}

In response to the complexities surrounding the  question of whether a given dataset is ``synthetic'' or not, we propose a conceptual framework whose aim is to provide structure and clarity when describing datasets that have a synthetic component, and to facilitate meaningful comparisons.
For our purposes, we define a \emph{dataset} as a collection of information-bearing objects that have been purposefully assembled or otherwise grouped, with the goal of serving a particular representational function, with the intention of being used in analysis (broadly defined) of some kind.
See \autoref{subsec:what_is_dataset} for a detailed discussion of this definition and its constituent elements.

How does our above definition fit with \textit{synthetic} data?
A dataset is, by definition, a human-generated artifact. 
At all stages of collection, curation, and processing, choices are made that separate the resulting dataset from its originating context.
As such, we further define a ``naturalistic'' dataset as one that fulfills the terms of our above definition, along with the additional constraint that the microdata contained in the dataset are \textbf{a)} intended by the designers to be \textit{faithful representations of some genuine originating phenomenon}, and \textbf{b)} are understood to be such by the users of the dataset.

For example, consider the MedDialog corpus of patient-provider dialogues taken from an online medical consultation website~\citep{zeng-etal-2020-meddialog}. 
The designers of this dataset understood and believed these dialogues to represent exchanges that actually took place between human patients and human providers who actually existed.
While they may have applied criteria to which dialogues were included (perhaps focusing on a particular category of patient, or clinical specialty), the lexical, semantic, pragmatic properties of the language and sociolinguistic aspects of the clinical dialogue were left un-modified in the hope of comprising a maximally representative sample.
In turn, downstream \textit{users} of this dataset would understand its contents to be representative of exchanges that actually took place (i.e., to be ``real messages''). 

In comparison, consider the \texttt{ACI} subset of the ACI-bench dataset~\citep{yim2023aci}, which consists of transcribed exchanges between an actual doctor and a volunteer ``patient,'' acting out a scenario based on symptom-driven prompts. 
The exchanges in such a dataset are ``real'' in the sense that they did, in fact, take place, and occurred between actual humans. 
However, they are not ``real'' in that they arose from an artificial scenario, as opposed to the actual real-world domain that they are attempting to represent.

The creators of ACI-bench, of course, attempted to design their scenarios to be as ``realistic'' as possible, and gave careful consideration to their design. 
Their goal was to create a dataset whose synthetic nature did not affect those aspects of its form or content to an extent that compromised the dataset's fundamental \textit{validity}\footnote{Here we use ``validity'' in the specific senses of construct and measurement validity; see \citet{brewer_crano_2014} for an overview of these terms, and \citet{jacobs_wallach_2021} for a discussion of how these concepts apply in machine learning research.} for the analyses of the sort that the dataset was intended by its authors and users to support.
Decisions about whether, when, and how to \emph{use} such a dataset typically hinge on the extent to which this assumption holds for one's particular purpose.

Finally, consider the DoPaCo dataset~ \citep{chen2023investigating}.
Here, the authors began with transcribed and de-identified genuine doctor-patient conversations, to which they had extremely restricted and time-limited access, and which could not be re-distributed. 
They used this corpus to fine-tune a pre-trained dialogue-generation Transformer model, and then used this model to generate a large corpus of synthetic dialogues, which they used for the rest of their work.
One may argue that such a dataset is in some sense \textit{meaningfully or usefully representative} of a ``real'' dataset of doctor-patient conversations, according to any number of rubrics.\footnote{Given that it was produced via a sequence-to-sequence model, one may accurately and literally describe it as consisting of a set of samples drawn from a statistical model of the originating ``real'' dataset; as such, there exist statistical measures that we may use to quantify the degree to which such samples are or are not representative (in a purely statistical sense) of the originating data, such as perplexity and KL-Divergence. However, we also note that there is a long-standing and robust literature~\citep{Galliers:1993,iyer_1997,ito_1999} on the \textit{limitations} of such purely statistical measures in terms of their external and ecological validity~\citep{brewer_crano_2014,Egami:2023}.}
But, it is also equally true that such a dataset, however \textit{statistically} similar it may be to its original training data, is not comprised of ``real'' dialogue data in the same sense as  ACI-bench or MedDialog.
Rather than representing true discursive choices made by humans interacting with one another, the dialogues in the DoPaCo dataset reflect that model's statistical approximation of the discursive choices represented in its  training data.

This question of ``real-ness'' is an entirely separate question from whether such a dataset is \textit{useful} or \textit{sufficient} for a given purpose. We are again using ``real'' as a \textit{descriptive} rather than a \textit{normative} term, and of course even highly synthetic datasets have many uses.
As such, we are emphatically \textit{not} arguing that datasets that are more naturalistic are inherently and categorically superior in some way to datasets that are more synthetic, though there may indeed exist scenarios where this is the case; that question is orthogonal to our purpose in the present work.
Finally, we note that the synthetic datasets we describe in this paper are uniformly accompanied by more or less detailed evaluations by their creators, seeking to investigate the degree to which they are or are not sufficient for specific applications.

\subsection{A Typology for Human/Machine Microdata Interventions}
\label{subsec:typology}

We propose a three-level typology of ``types'' of intentional intervention to the \textit{content} of a dataset's microdata (illustrated in \autoref{tab:typology}). Our typology's categories are inspired by the perturbation-generation dichotomy described by~\citet{domingo-ferrerSurveyInferenceControl2008}, adapted and expanded in terms appropriate to the complexities of linguistic datasets (as opposed to traditional statistical datasets). 
Additionally, we observe that the interventions involved in producing a synthetic linguistic dataset may be carried out by either humans or machines (or both) during the course of a dataset's creation. 
In practice, many synthetic datasets involve a mix of both, and adequately capturing the ways in which a given dataset is ``synthetic'' requires that both aspects be described. 
As such, our typology involves categorizing a dataset along dimensions of both the \textit{human} and \textit{machine} involvement~(\autoref{tab:typology}).
This increased expressiveness further helps break down the false binary of ``synthetic'' vs. ``real,'' and allows clearer descriptions of a given dataset.


\begin{itemize}
\item Type 1. No intervention (i.e., there is no automatic or manual process of alteration, and data reflect naturally-occurring linguistic exchanges).
\item Type 2. Existing microdata is perturbed according to a clear specification (e.g., anonymization, paraphrasing, redaction). These perturbations are traceable, with the synthetic data maintaining a direct lineage to the source data. Human- or machine-mediated feedback may guide these alterations, but the structure of the source data is preserved to a recognizable degree.
\item Type 3. New microdata are produced \textit{de novo} via a generative process of some kind, to substitute for ``real'' microdata. The generation may be conditioned on real data, and may also involve exogenously-provided knowledge in the form of prompts or templates. 
Iterative refinement involving human feedback or reinforcement learning (e.g., via RLHF\footnote{Reinforcement Learning from Human Feedback~\citep{christianoEtAl2017RLHF}.} or human simulations) falls under this category when it results in the creation of fundamentally new data samples.
\end{itemize}

In our typology, these levels are not mutually exclusive. It is possible for a dataset to include both Type 2 and  Type 3 interventions. 
Similarly, a dataset may involve multiple interventions of the same or a different type, focused on different aspects of its contents.


Consider a dataset of ``screen-written'' patient-provider interactions, in which the dialogues' contents and discursive structure are entirely imagined by humans in response to a prose case description (for example, the MTS-Dialog dataset~\citep{ben-abacha-etal-2023-empirical}). 
As the representational goal of this dataset is to contain ``realistic'' clinical dialogue, to be used in place of otherwise unavailable genuine dialogues, we consider this to be a ``Human Type 3, Machine Type 1'' dataset.

However, 
if the \textit{goal} of the dataset had been to consist of simulated dialogues (as opposed to the actual goal, to wit, using \textit{simulated} dialogues as \textit{representative stand-ins} for \textit{genuine} dialogues), we would instead consider the human-written dialogues to be non-synthetic primary microdata, and categorize the human component of this dataset under ``Type 1.''

This example highlights the value of our typology. It facilitates clarity about both the representational goals and the actual content of a given dataset. By doing so, we make it easier for future producers of datasets to communicate explicitly about these aspects of their creations, and for consumers of datasets to reason concretely about whether their intended use aligns with the dataset's capabilities and contents.



\subsection{In Search of Synthetic Dialogues}

Our typology's design was informed by a literature review whose goal was to survey the landscape of synthetic biomedical dialogue datasets.
We defined our clinical domain of  interest as comprising text capturing either spoken or written language  generated by patients and/or health providers, intended for use as part  of clinical care, patient information-seeking, etc. 
Our goal was to identify published datasets of clinical dialogues, in spoken or written genre, and either synthetic or natural in nature.

Included publications could be specifically focused on the production and evaluation of their datasets, or could produce datasets as secondary research artifacts. 
In the latter case, we required that there be an intention on the part of the authors that their datasets be re-used by others (i.e., their data were not simply made re-usable for purposes of replicability). 
We used keyword-based and index-term search strategies, supplemented by large language models (LLMs) to assist in filtering and classifying retrieved papers. 
See \autoref{sec:appendix-search} for a detailed description of our approach and \autoref{sec:appendix-prompts} for the verbatim LLM prompts. \autoref{tab:search-summary-numeric} summarizes the results of our structured search strategies across PubMed, the ACL Anthology, and dblp.

From a total of 1,626 papers (679 from PubMed, 591 from ACL Anthology, and 356 from dblp), successive filtering steps (including LLM-assisted classification and manual review) reduced the pool to 25 included papers (4 from PubMed, 11 from ACL Anthology, and 10 from dblp). After removing overlapping entries, we arrived at 20 unique papers.
 

From a scientific standpoint, we note that clinical dialogue-oriented datasets are more variable in terms of their linguistic domains, intended uses, structure, means of collection, methods of synthesis, etc. than the larger but comparatively more homogeneous body of literature describing synthetic electronic health record (EHR) notes.
As such, they provide an informative corpus of examples to use in attempting to understand and categorize different approaches to synthesis.
In practice, however, our typology (and this paper's broader discussion of synthesis) would be applicable to datasets containing synthetic notes or other forms of clinical language.

Given our focus on dialogues, we excluded the following categories of dataset from our analyses in this paper:

1. Published datasets whose primary content was medical in nature but did \textit{not} consist of dialogues, e.g.,  electronic health record (EHR) notes, structured EHR data, etc. This resulted in the exclusion of most existing work on data augmentation and synthetic medical data more generally, and on clinical language in particular. By far the most frequently-seen category of synthetic clinical language dataset is that of EHR notes.

2. Datasets where the synthetic component was in the form of structured annotations (e.g., dialogue act classification, concept/entity labels, etc.), narrative summaries, or other secondary metadata, as opposed to the primary dialogue itself.

3. Datasets that \textit{use} genuine dialogues, but whose artifacts of interest and analysis are synthetically-produced secondary products (such as summaries or clinical notes) generated automatically from those original genuine dialogues.

We also excluded datasets that are not themselves presenting or simulating dialogues as such, but which \textit{are} intended to produce artifacts for use in training systems whose modality of use is conversational in nature. 
For example, \citet{kang-etal-2024-self} describe a process for transforming curated biomedical data into pseudo-conversational question-answer pairs for use in instruction tuning a downstream LLM. 
While the intended \textit{purpose} of such a system would be to support conversational interaction between a human and an LLM, \textit{the data itself} are not conversational in nature nor does it attempt to characterize the key linguistic features of a dialogue-based interaction. 

As such, papers describing such datasets were of secondary interest, and we have excluded them from our review. We note that, alongside synthetic EHR notes, synthetic QA pairs are also a very prevalent form of synthetic clinical dataset.

Another category that we ultimately excluded from our review are papers that generated synthetic dialogue data as a step towards some \textit{other} primary analysis, but which did not release their resulting datasets as artifacts for use by others.
One prototypical example of this category of work includes that of \citet{yosef2024assessing}, who carried out a series of experiments designed to assess the ability of LLMs to simulate patient/therapist interactions.
While they do produce synthetic dyadic interactions as part of their work, their primary interest is using them as inputs to another LLM-based analysis designed to determine the ability of LLMs to differentiate between styles and forms of therapy.
A second example of this category of work would be that of \citet{grayIncreasingRealismVariety2024}, who used LLMs to simulate patient-provider counseling discussions in a neonatology context and conducted a very thorough analysis in order to assess their content and utility from a clinical perspective.

In both cases, the authors did not publish their resulting simulated interactions, and in both cases, producing these simulated exchanges was not the primary (or even secondary) point of their study.
Notably, if they \textit{had} published their linguistic datasets, the resulting corpora from both papers would be usefully described by our typology (both would be Human-1/Machine-3, in that all linguistic microdata were generated entirely ``from scratch'' by the machine with no human involvement). 

\begin{figure}
\hspace*{-.65cm}  \includegraphics[width=1.1\columnwidth]{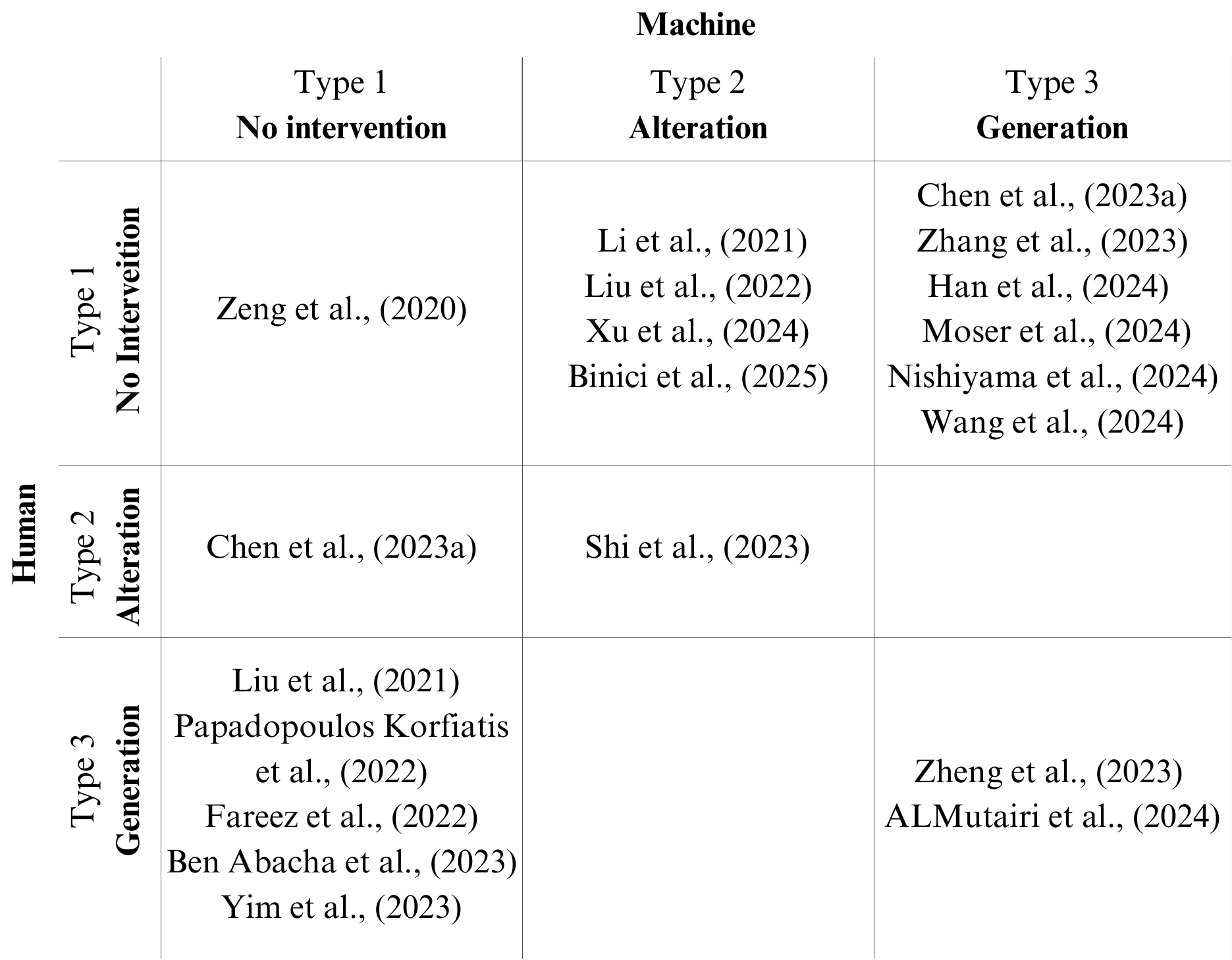}
  \caption{A selection of clinical dialogue datasets classified according to different degrees of human-machine interventions.
  }
  \label{fig:mainfigure}
\end{figure}




\section{Discussion}

\subsection {Synthetic Data and Varieties of Task}
\label{subsec:synth_data_variety}

As discovered in our review, synthetic datasets in the clinical domain are used for a variety of NLP related tasks such as: Information Extraction \citep{zhang-etal-2020-mie}, summarization of the clinical dialogues \citep{joshi-etal-2020-dr,chintagunta-etal-2021-medically}, generating notes from conversations \cite{yim-etal-2020-alignment}, dialogue with custom personality traits for training purposes \cite{han-etal-2024-psydial-personality}, training data for ASR systems \cite{fareez2022dataset}, and named entity recognition \cite{frei2023annotated}.

\textbf{Language Diversity in Datasets:} As is commonly observed across NLP tasks, English was the most prominent language among the datasets we identified (e.g. \citet{joshi-etal-2020-dr,chintagunta-etal-2021-medically,abercrombie-rieser-2022-risk,wang-etal-2023-umass,ben-abacha-etal-2023-empirical,yim2023aci}). In addition, there is a growing amount of research for Chinese (Mandarin) datasets \cite{zhang-etal-2020-mie,shi-etal-2023-midmed} as well as in Arabic \cite{almutairi-etal-2024-synthetic}, Korean \citep{han-etal-2024-psydial-personality}, and German \citep{rohrig2022grascco,frei2023annotated}.

\textbf{Type of Medical Domains:} Most datasets we were able to retrieve are not targeted for a specific type of disease in the clinical domain, and instead attempted to more broadly represent written clinical notes or patient-provider dialogues from generic clinical settings (e.g. an outpatient visit). 
However, we did encounter a handful of synthetic datasets focusing on specific clinical specialties, including cardiology \citep{zhang-etal-2020-mie}, gastroenterology \citep{liu2022meddg}, psychology \citep{liu-etal-2021-towards,zheng-etal-2023-augesc,han-etal-2024-psydial-personality}, and obstetrics \& gynecology \citep{Wang:2023d}.

\subsection{On Beyond Microdata}
\label{subsec:beyond_microdata}

To this point, we have limited our discussion to the lowest and most granular level of a dataset's contents, its microdata.
Beyond what we have presently described, there is also another dimension of synthesis that we believe to be relevant.
A linguistic dataset includes information at multiple levels of representation, including lexical (which words are used), syntactic (how are sentences constructed), and semantic (what meaning is being expressed).
In the case of a dialogue-oriented dataset, there may also be pragmatic and discursive information (who is speaking, in what register(s) and with what turn-taking structure, etc.).
Additionally, there may be extra-linguistic information represented in such a dataset, for example, the larger social and phenomenological context from which the data are meant to have arisen. 
Interventions or other synthetic data generation processes may be targeted at, or otherwise affect, all, some, or none of these layers.

There are also other factors that play important roles in the ``realism'' of a dataset.
Recall the MTS-Dialog dataset of \citet{ben-abacha-etal-2023-empirical}, which consists of ``dialogues'' that were imagined and ``screen-written'' \emph{de novo} by humans (in this case, trained clinicians) based on short clinical case vignettes.
Now, imagine a hypothetical second dataset (``dataset \textit{B}'') in which, instead of being screen-written dialogue, the discourse samples were in fact the result of transcribed conversations taking place between two actors, posing as a doctor-patient dyad and improvisationally acting out a scenario based on the same case vignettes; the dataset described by \citet{fareez2022dataset} follows this method of creation, as does the \texttt{ACI} subset of the ACI-bench dataset~\citep{yim2023aci}.

In both of these examples, we would consider the resulting dataset to be synthetic, in the sense that neither one reflects the language produced by actual ``real world'' participants in the true interaction that the dataset is purporting to represent (i.e., a clinical encounter).
Our typology in its present form would categorize both as Type 3 datasets along their human dimensions,\footnote{As the microdata are being ``made up'' (as opposed to having naturally arisen or being modified \emph{in situ}).} and in both cases we can safely say that the context that gave rise to the linguistic information contained in the dataset is surely more artificial than in a hypothetical \textit{third} dataset (``dataset \textit{C}'') containing ``genuine'' conversations between patients and providers captured ``in the wild'' (a Type 1 dataset, according to our typology).

However, this thought experiment illustrates a limitation of our present typology.
The linguistic artifacts in ``dataset B'' (e.g., the utterances themselves, along with their associated paralinguistic and discursive content) were produced by actual humans in conversation, and could certainly be thought of as being ``less synthetic'' than the utterances in MTS-Dialog, which were made up from whole cloth.
In both scenarios, the semantic content of the utterances may be quite similar, as might the degree to which they accurately reflect the vignettes that formed their seeds. 
The variation would involve the above-mentioned pragmatic and discursive aspects of the data, as a side effect of the difference in the method of generation.
Our typology currently lacks a notion of, for lack of a better word, \textit{modality}: in this case, ``imagined'' vs. ``transcribed'' spoken language. 
It also does not distinguish between degrees of synthesis along these dimensions.

Another area that our typology does not capture is the relationship between synthesis and a dataset's sociolinguistic context \cite{nguyen-etal-2016-survey}.
Consider a hypothetical dataset of transcribed English-language patient-provider conversations taking place during genuine outpatient visits in the United States (i.e., a Type 1, ``real'' dataset in our typology). 
Imagine, now, that the dataset were run through a machine translation (MT) algorithm, with the intent of producing a French-language dataset for use in evaluating algorithms designed to process Francophone clinical encounters.
According to our typology, this is now a ``Type 2'' dataset: the dataset's microdata are still clearly derived from ``real'' microdata, but have been perturbed in some systematic way.

Depending on the quality of the machine translation system, the resulting dataset may be ``equivalent'' in terms of its lexical and syntactic representativeness (due to the similarity between English and French), and may also be equivalently useful in terms of its semantic content (i.e., the set of clinical conditions and events being described).
In other words, the impact of the synthesis may be minimal along those dimensions.
However, the \textit{pragmatic} and \textit{discursive} content may be inappropriate, as norms of clinical communication and patient expectations vary significantly between cultures and countries.\footnote{We note that, even within societies, norms of clinical communication vary greatly across numerous social and demographic factors, as well as  between clinical specialties. This is one of the reasons why we are particularly wary of synthetic data in the context of clinical dialogues.}
As such, even assuming a very high-quality translation, the dataset is likely no longer as valid, in terms of its linguistic representativeness, as it may have been before being made synthetic. 
This is likely the case to a degree that is much larger than might be expected by a linguistically naïve researcher equipped with an MT algorithm.\footnote{Despite technical advances, translation remains, of course, notoriously challenging~\citep{Anonymous:1968p9217} and finicky~\citep{Carlson:2007p3777}. Clinical contexts bring numerous specialized challenges for the use of MT~\citep{mehandru_2022}. }

Additionally, consider that the extra-linguistic factors represented in our hypothetical dataset may be severely impacted by this sort of perturbation.
In this particular example, the translated conversation may indeed be adequately representative in terms of its linguistic and clinical content, but its contextual representativeness may be impaired.
A patient-provider discussion about the management of  end-stage kidney disease in American English will include discussion of lab values, diet, etc., all of which may translate into a French context with adequate fidelity. 
However, the parts of the conversation involving the logistics of insurance coverage for dialysis, or of organizing a social worker to assist in transportation to and from dialysis sessions, would differ wildly between the two contexts.

This scenario is inspired by an actual example we encountered during our literature review, in which ACI-bench (a Human Synthetic Type 3 dataset) was used in a one-shot setting to prompt LLMs to generate equivalent dialogues in the Najdi Arabic dialect, with the goal of creating a dataset for use in cultural adaptation of an LLM~\cite{almutairi-etal-2024-synthetic}. 
According to our typology, the resulting data is machine-generated while also being originated by a human-driven \textit{de novo} synthesis process and falls under Human Type 3, Machine Type 3. 
As LLMs are increasingly deployed at scale, hybrid synthetic datasets of this type will likely become more common. 
However, the validity and representativeness of such datasets must be carefully assessed.

\section{Conclusion}
\label{sec:conclusion}
Although synthetic datasets are widely used for dialogue generation, the boundaries of what makes a dataset synthetic are not always clear, and they vary among fields. Focusing on the healthcare domain and only on dialogue data, we propose a typology to clarify the types of synthetic datasets in terms of the human and machine interventions.

Academic literature on synthetic data focuses heavily on the mechanics of \textit{how} synthetic data elements are generated, and our typology focuses instead on the role that the synthesis plays in the overall ``shape'' of the dataset.
Our typology provides authors with the ability to speak in a more clear and granular way about what aspects of their dataset are synthetic, to what degree, and from what source (human vs. machine).

Our literature review revealed a clear pattern of increased prevalence of fully-synthetic (Type 3) datasets, particularly of machine-generated dialogues following the widespread availability of LLMs in 2023.
Given this, our work's broader perspective on syntheticity will help researchers better assess and contextualize the validity and representativeness of these datasets.

The relationship between synthesis and a dataset's larger contextual factors remains currently unexplored. 
As future work, we hope to expand our work to capture realism and synthesis along non-linguistic and contextual factors.\footnote{The extent to which there exists a valid dichotomy between ``linguistic'' and ``non-linguistic'' content in a dataset is surely up for debate. }

\section{Limitations}
Our typology provides a structured framework for describing synthetic dialogue datasets and we summarize here some of its limitations discussed broadly in
\S\ref{subsec:beyond_microdata}.

First, the categories of microdata interventions do not capture variations in modality (i.e., whether dialogues are imagined by a single author, simulated through human interaction, or transcribed from naturalistic speech). As such, pragmatic and discursive elements may differ significantly between these modalities.

Second, our work does not account for the sociolinguistic context of the synthetic data. A dataset may be linguistically equivalent after MT, yet fail to reflect cultural norms of clinical communication, potentially reducing its validity.

Currently, the ability to characterize amounts of synthesis along these dimensions is limited, and at the present time, we consider these questions to be left for future work. 



\section*{Acknowledgments}
Sergiu Nisioi is supported by the project “Romanian Hub for Artificial Intelligence - HRIA”, Smart Growth, Digitization and Financial Instruments Program, 2021-2027, MySMIS no. 351416. Steven Bedrick is supported by the National Library of Medicine of the National Institutes of Health under award number 5R01LM011934-10 (``Semi-structured Information Retrieval in Clinical Text for Cohort Identification'').

\section{Bibliographical References}\label{sec:reference}

\bibliographystyle{lrec2026-natbib}
\bibliography{custom,additional_refs}


\iftrue
\appendix

\section{What is a Dataset?}
\label{subsec:what_is_dataset}

In response to the complexities surrounding the  question of whether a given dataset is ``synthetic'' or not, we propose a conceptual framework whose aim is to provide structure and clarity when describing datasets that have a synthetic component. To facilitate meaningful comparisons, we must first define what we mean by ``dataset,'' and propose here a broad and ``teleological'' definition.
We set as a foundation that a dataset is a finite collection of information-bearing objects, in the sense described by \citet{bucklandInformationThing1991}. Such a collection may be either digital or physical in nature.
Note that we refer here to a ``collection'' as opposed to a ``grouping'', ``assembly'', ``set'', etc.
Not all groupings of information objects constitute datasets. A random and incidental set of bits would not constitute a dataset, nor would a random pile of books encountered on the roadside.

A \textit{collection} implies a \textit{collector} (i.e., an external agent with at least some degree of intentionality behind its assembly). Furthermore, a \textit{collection} necessarily brings with it some notion of an intended \textit{user} whose needs or goals the collection is meant to support.\footnote{The user and the collector may, of course, be the same entity.}
Another key property of a \textit{collection} is that it involves some kind of extrinsic criteria for what is and is not included in the collection.\footnote{In the context of traditional library and archival collections, questions of ownership and accessibility also play into the definition of a collection~\citep{Lee:2000}.}

This definition, however, is insufficient. An intentionally collected set of bits representing random numbers may not \textit{on its own} constitute a dataset.
Therefore, we introduce a second constraint to our definition. 
Our collection of information-bearing objects must be \textit{documents} in the sense meant by \citet{briet1951}, who defined a document as ``any concrete or symbolic indication, preserved or recorded, for reconstructing or for proving a phenomenon, whether physical or mental.''\footnote{``Tout indice concret ou symbolique, conservé ou enregistré, aux fins de représenter ou de prouver un phénomène ou physique ou intellectual.'', \citet[p. 7]{briet1951} cited in \citet{bucklandInformationThing1991}.}

According to this definition, a crucial distinction between ``a collection of random bits'' and a ``dataset'' is that the latter has an underlying \textit{intention} of some sort on the part of whoever is identifying it as such (its creator or collector), and that this intention is reflected in the nature of the dataset's contents. 
The ``concrete or symbolic indication'' is being intentionally ``preserved or recorded'' \textit{for} some purpose. Thus, we can conclude that it is necessary for that purpose that it be a \textit{meaningful} and \textit{representative} representation. Otherwise it would be useless for ``reconstructing or ... proving a phenomenon.''\footnote{In the case of a dataset comprising random noise, the intended point of the dataset is that the noise is in fact random according to some particular definition or scheme, and the existence of the dataset would constitute an assertion of that property on the part the dataset's authors.} 

Under this part of our definition, we argue that part of what makes a dataset a dataset is that it is \textit{intended} by its originators to be representative of \textit{``something''}, hence, our definition's description as ``teleological.'' 
What that ``something'' \textit{is} may vary wildly, as may the degree to which a given dataset actually represents whatever it is meant to represent. These are important questions, but even a dataset that is flawed, incomplete, broken, etc., still counts as \textit{a} dataset.

To illustrate this distinction, consider the bits representing the set of digital music files on one of the authors' laptops.
This is comprised of a large grouping of binary files and associated metadata, but taken on its own, stripped of context and intention, it is simply that: a grouping.
It serves no purpose and has no intrinsic meaning, other than to exist and be stored in his computer's filesystem. Its constituent bits may be interpretable by this or that piece of software, but are essentially the digital equivalent of a random and incidental pile of books. 

In this particular case, it is reasonable to argue that they represent a \textit{collection}, in the sense previously described. They have been intentionally collected, in that they have been purposefully assembled over a period of time as new music has been added, and there also exist numerous possible digital music files that are \textit{not} in the collection.
However, it would be nonsensical to say that this particular collection is or is not ``representative'', ``meaningful'', ``useful'', ``complete'', ``well-formed'', etc.: each of those descriptors implies a \textit{telos} of some kind. In other words, each immediately prompts the questions, ``for what?'' and ``for whom?'' ``Meaningful'' --- to whom, in what way, and in what context? The files in their present form lack any such thing --- they are simply files.

Now, imagine that the author in question decides to try and \textit{use} this collection as part of an analysis, perhaps to visualize his music consumption patterns.
We argue that it is at this point, and not before, that our collection of information objects becomes a \textit{dataset}: the point at which it is first intentionally collected \textit{in order to represent, reconstruct, prove, etc. a particular phenomenon}.
In other words, the point at which the bits become part of a process of problem formulation.\footnote{See \citet{Passi:2019} for an excellent discussion on this topic, including on the contingent and bidirectional relationship between data and problem formulation.}

Next, we add an additional constraint: in addition to the intention on the part of somebody to use a collection for some particular purpose, we also argue that there must also be an intention for some other party to understand the collection as being similarly meaningful. 
This is closely linked to the notion of ``audience'' in our discussion of collections above, but is more specifically linked to the \textit{telos} of the dataset in question.

At this point, our definition has a ``what'' (collection of information objects), a ``when'' (the point at which somebody treats the collection as being a meaningful representation of a particular phenomenon), and a ``who'' (whatever additional audience or user community the author expects to align with them as to the nature and purpose of the dataset). 
In the interests of theoretical completeness, we add one more constraint, which may be categorized under the heading of ``whether,''\footnote{In the descriptive sense of ``whether or not it is possible'', not the prescriptive sense of ``whether or not it is a good idea''.} and argue that for a collection to be considered a dataset, it must also contain information objects that in terms of their structure and content could plausibly be used from a functional or operational standpoint to achieve the dataset's purpose.

In other words, if the point of the dataset is to visualize the author's music consumption patterns, it must necessarily be comprised of documents containing the appropriate kinds of information (e.g., digital files containing music, metadata recording listening patterns, financial records of purchases, etc.), as opposed to being (e.g.) made up of random bits, pictures of cats, etc.
This part of our definition may at first appear somewhat tautological. However, we include it to help rule out counter-examples and clarify the functional nature of our definition.\footnote{We express our thanks to Dr. Colin Koopman for this part of our definition~\citep{koopman_2025}.}
An analogy with measurement theory~\citep{brewer_crano_2014} is appropriate. This part of our definition attempts to rhyme more with ``face validity'' than it does with ``construct validity'' or ``measurement validity'', in that we are not referring to how \textit{well} the datasets' contents serve its intended purpose but rather are speaking about basic functional plausibility.

Nothing in our definition means that a dataset built to serve one \textit{telos} may not be used in pursuit of another, or that a dataset may not be designed in pursuit of multiple \textit{telea}. It does, however, help make apparent the fact that re-purposing of data is a less straightforward undertaking than it often appears and requires careful consideration.
Additionally, our definition has nothing to say about what \textit{kinds} of purposes are or are not ``valid,'' or limits to how general or specific the purpose of a dataset may be.

\section{Search Strategies}
\label{sec:appendix-search}
\subsection{PubMed Search Strategy}
\label{sec:appendix-pubmed}

Our PubMed search strategy consisted of three separate searches, with the goal of identifying published datasets of clinical dialogues, either synthetic or natural.
This was conducted as a supplement to our preliminary literature search conducted using less precise methods, and should be considered current as of April, 2025.

The first search used solely keywords, and attempted to directly capture papers reporting the creation of datasets relating to dialogues, and was as follows: \texttt{
(((patient AND provider) or (clinical)) AND (conversation* OR dialog OR dialogue)) AND (corpus OR dataset)
}

The second attempted to leverage MeSH index terms, and did not specifically include terms relating to datasets or corpora, but did add a handful of keywords to reduce scope: \texttt{
("Artificial Intelligence"[MeSH Terms] OR "Machine Learning"[MeSH Terms] OR  "Natural Language Processing"[MeSH Terms] OR "Information Storage and Retrieval"[MeSH Terms] OR "Pattern Recognition, Automated"[MeSH Terms]) AND ("Physician-Patient Relations"[MeSH Terms] OR "Referral and Consultation"[MeSH Terms] OR "Interviews as Topic"[MeSH Terms] OR "Patient Simulation"[MeSH Terms] OR "Communication"[MeSH Terms]) AND ("conversation" OR "dialogue" OR "dialog" OR "discourse")}

The third returned to a keyword approach, and focused on patient-provider conversations: \texttt{
(((patient AND provider) or (clinical)) AND (conversation* OR dialog OR dialogue)) AND (corpus OR dataset)}

The three searches together yielded 679 unique results, of which the vast majority were false positives. 
An LLM (Gemma 3 21b) was used to perform further filtration.
First, we instructed the model to classify each retrieved paper according to whether its abstract described ``automated processing of either spoken or written clinical language in the context of patient-provider communication.''
This reduced the set to 371 articles; manual inspection revealed satisfactory performance, and we were able to verify that specific known true-positive results were correctly classified by this method, further supporting its efficacy.

We, then, performed a second step of LLM-based filtration, instructing the model to classify each of those articles as to whether the abstract was ``describing a paper that created a new corpus or dataset with the intention of it being re-usable by others.'' The model was instructed to prioritize recall over precision.

This reduced our set of articles to review to 125. Both prompts are rendered in the boxes from \autoref{sec:appendix-prompts}.
We used similar heuristic methods to verify that this filtering method performed at a satisfactory level of accuracy. The precision was reasonably high, in that articles in this set did in general contain notable discussion of the details of the datasets that they used, and generally did all describe the creation of novel corpora.
However, many were describing corpora that were outside of our review's scope (i.e., datasets of QA pairs, etc.), or described corpora that were not releasable due to reasons of privacy or other such constraints. Only 4 papers matched our criteria and were included in our review.


\subsection{ACL Anthology Search Strategy}
\label{sec:appendix-acl}
We implemented a script to retrieve, filter, and annotate papers from the ACL Anthology \url{https://aclanthology.org/}. The process involves scraping the HTML content of the full ACL proceedings between 2019-April, 2025 (EMNLP, ACL, EACL, NAACL, LREC) including all the workshops associated. Each paper title, abstract, authors and paper URL was extracted.

Papers were filtered based on the presence of the keyword \texttt{medic} in either the title or the abstract. This filtering is high recall, case-insensitive, and matches partial substrings, resulting a total of 591 papers. 
For each filtered paper, the corresponding full text document is extracted and submitted to the ChatPDF API.\footnote{\url{https://www.chatpdf.com/}} A fixed prompt (see \autoref{sec:appendix-prompts}) was used to request a structured JSON response with answers to the following binary and list-format questions: (1) whether a synthetic dataset was created, (2) whether the dataset contains dialogues, (3) whether human evaluation was conducted, (4) whether the dataset was used in a downstream task. We further reduce the list to 180 papers based on criteria (1) and (2). A manual review of the titles and abstracts further narrowed the set to 23 papers. Full-text examination of these 23 papers resulted in 11 that matched all criteria of interest and were included in our final review.

\subsection{dblp Search Strategy}
\label{sec:appendix-dblp}

The search employed on \url{https://dblp.org} was broad and covered a range of related fields, including speech processing, medical informatics, dialogue systems, artificial intelligence, argumentation, and information retrieval. To ensure high recall, \footnote{\texttt{(med* | syn* | sim*) \& dial*}} the search targeted titles containing words beginning with \texttt{med*} (e.g., medical, medication), \texttt{syn*} (e.g., synthesis, synthetic), or \texttt{sim*} (e.g., simulation, simulating) in combination with \texttt{dialo*} (e.g., dialogue).

To refine the results, a filter was applied to exclude publications from major venues associated with the Association for Computational Linguistics (ACL) and preprints. The final corpus was restricted to articles published in selected conferences, workshops, and journal series,\footnote{In total 362 \texttt{streamid's} corresponding to different publications.} including \textit{INTERSPEECH}, \textit{EUROSPEECH}, \textit{ICASSP}, \textit{IJCAI}, \textit{SIGIR}, \textit{AAAI}, \textit{MEDINFO}, the \textit{Journal of Biomedical Informatics}, \textit{BMC Medical Informatics and Decision Making}, \textit{Computers in Biology and Medicine}, \textit{CEUR-WS}, \textit{FAIA}, \textit{CCIS}, \textit{SCI}, and relevant Springer proceedings, among others.

In total, 356 papers were identified through this search. 
Based on title screening, 45 of the retrieved papers were selected for further manual review of abstract and content. Among these, only 10 papers have been marked as suitable for our review. 

\clearpage
\onecolumn

\section{Synthetic Datasets}
\label{sec:appendix}

\begin{table*}[ht]
\centering
\resizebox{0.99\textwidth}{!}{%
\begin{tabular}{l|c|c|c|c}
\textbf{Source} & \textbf{Initial Hits} & \textbf{After First Filter} & \textbf{After Review} & \textbf{Included} \\
\hline
PubMed & 679 & 371 (LLM: clinical language) & 125 (LLM: dataset creation) & 4 \\
\hline
ACL Anthology & 591 & 180 (LLM: dataset+dialogue) & 23 & 11 \\
\hline
dblp & 356 & - & 45 & 10 \\
\end{tabular}%
}
\caption{Search strategy summary. The first filter for PubMed and ACL Anthology are based on LLM prompting. There is an overlap between the three search strategies and the total number of unique papers is 20.}
\label{tab:search-summary-numeric}
\end{table*}

\begin{table*}[htb]
\centering
\resizebox{0.99\textwidth}{!}{%
\begin{tabular}{llll}
\textbf{Dataset}                                                              & \textbf{Acronym}    & \textbf{Language} & \textbf{Types}                                                                  \\
\hline
\citet{ben-abacha-etal-2023-empirical}             & MTS-DIALOG & en       & \begin{tabular}[c]{@{}l@{}}Human Type 3, Machine Type 1
\end{tabular}                      \\
\hline
\citet{liu-etal-2021-towards}                      & ESConv     & en       & \begin{tabular}[c]{@{}l@{}}Human Type 3, Machine Type 1
\end{tabular}                         \\
\hline
\citet{fareez2022dataset}                          &           & en       & \begin{tabular}[c]{@{}l@{}}Human Type 3, Machine Type 2\end{tabular} \\
\hline
\citet{papadopoulos-korfiatis-etal-2022-primock57} & PriMock57  & en       & \begin{tabular}[c]{@{}l@{}}Human Type 3, Machine Type 2\end{tabular}     \\
\hline
\citet{yim2023aci}                                 & ACI-bench  & en       & \begin{tabular}[c]{@{}l@{}}Human Type 3, Machine Type 2\end{tabular}            \\
\hline
\citet{almutairi-etal-2024-synthetic}              &            & ar-sa    & \begin{tabular}[c]{@{}l@{}}Human Type 3, Machine Type 3\end{tabular}\\
\hline
\citet{zheng-etal-2023-augesc}                     & AugESC     & en       & \begin{tabular}[c]{@{}l@{}}Human Type 3, Machine Type 3
\end{tabular}                          \\
\hline
\citet{Chen:2023c}                                 & IMCS-21    & zh       & \begin{tabular}[c]{@{}l@{}}Human Type 2, Machine Type 1\end{tabular}\\
\hline
\citet{shi-etal-2023-midmed}                       & MidMed     & zh       & \begin{tabular}[c]{@{}l@{}}Human Type 2, Machine Type 2
\end{tabular}   \\
\hline
\citet{zeng-etal-2020-meddialog}                   & MedDialog  & zh / en  & \begin{tabular}[c]{@{}l@{}}Human Type 1, Machine Type 1\end{tabular} \\
\hline
\citet{liu2022meddg}                               & MedDG      & zh       & \begin{tabular}[c]{@{}l@{}}Human Type 1, Machine Type 2\end{tabular}                                                           \\
\hline
\citet{li2021semi}                                 & KaMed      & zh       & \begin{tabular}[c]{@{}l@{}}Human Type 1,Machine Type 2\end{tabular} \\
\hline
\citet{xu-etal-2024-reasoning}                     & EMULATION   & zh       & \begin{tabular}[c]{@{}l@{}}Human Type 1, Machine Type 2\end{tabular}                             \\
\hline
\citet{binici2024medsage}                          & MEDSAGE    & ch       & \begin{tabular}[c]{@{}l@{}}Human Type 1, Machine Type 2\end{tabular}                            \\
\hline
\citet{zhang-etal-2023-huatuogpt}                  & HuatuoGPT  & zh       & \begin{tabular}[c]{@{}l@{}}Human Type 1, Machine Type 3\end{tabular}                                                         \\
\hline
\citet{nishiyama-etal-2024-assessing}              &            & ja       & \begin{tabular}[c]{@{}l@{}}Human Type 1, Machine Type 3\end{tabular}                                                         \\
\hline
\citet{han-etal-2024-psydial-personality}          & PSYDIAL    & ko       & \begin{tabular}[c]{@{}l@{}}Human Type 1, Machine Type 3\end{tabular}                                                         \\
\hline
\citet{chen2023investigating}                      & DoPaCo     & en       & \begin{tabular}[c]{@{}l@{}}Human Type 1, Machine Type 3\end{tabular}                                                        \\
\hline
\citet{wang-etal-2024-notechat}                    & NoteChat   & en       & \begin{tabular}[c]{@{}l@{}}Human Type 1, Machine Type 3\end{tabular}                                                         \\
\hline
\citet{moser2024synthetic}                         &            & en, de   & \begin{tabular}[c]{@{}l@{}}Human Type 1, Machine Type 3\end{tabular}      
\end{tabular}
}
\caption{The list of identified synthetic datasets and their categories.}
\label{tab:main}
\end{table*}

\section{LLM Prompts}
\label{sec:appendix-prompts}

\begin{tcolorbox}[colback=white!10, colframe=black, title=Gemma 3 27b: clinical language filter, label={box:gemma3-prompt2}, width=\textwidth]
Your role is to review an abstract of a journal article, and decide whether it should be included in a literature review. We are looking for articles about automated processing of either spoken or written clinical language in the context of patient-provider communication. Examples would include: clinical conversations, discourse, written narrative communications, etc. Either transcribed or recorded language is acceptable.

If you are not sure, or if the article is borderline, please consider it in-bounds and err on the side of inclusion.
\end{tcolorbox}

\begin{tcolorbox}[colback=white!10, colframe=black, title=Gemma 3 27b: dataset creation filter, label={box:gemma3-prompt1}, width=\textwidth]
Your role is to review an abstract of a journal article, and determine whether or not the abstract is describing a paper that created a new corpus or dataset with the intention of it being re-usable by others. Possible things to look for:

- The abstract might explicitly mention that the dataset produced as part of the article was released or otherwise made publicly available.

- The abstract might talk about how the authors developed a novel corpus, specifically because of a lack of already-existing data or because existing data was insufficient in some way.

Remember, all of these papers will have involved making a dataset or corpus of some kind. We are looking specifically for papers that where one of the final products of the research, or where the main point of the research, was to create a new dataset or corpus, especially if it is described as being released or made available.

Simply mentioning an already-existing dataset is not enough.
\end{tcolorbox}

\begin{tcolorbox}[colback=white!10, colframe=black, title=ChatPDF: dataset+dialogue, label={box:chatpdf-prompt}, width=\textwidth]
Provide a json response using the following structure:
\begin{lstlisting}[language=json,firstnumber=1]
{
"synthetic_dataset_created": true/false,
"contains_dialogues": true/false,
"human_evaluation": true/false,
"uses_dataset_in_downstream_task": true/false
}
\end{lstlisting}

To answer the following questions:

- Do the authors create a synthetic dataset in this paper?

- Does it contain dialogues?

- Does it have a human evaluation of the dataset?

- Does it use the dataset in a downstream task?

\end{tcolorbox}

\fi

\end{document}